\documentclass[10pt,twocolumn,letterpaper]{article}

\usepackage[pagenumbers]{cvpr} 

\usepackage{xcolor}
\usepackage{multirow}
\usepackage[T1]{fontenc}
\usepackage[utf8]{inputenc}
\definecolor{cvprblue}{rgb}{0.21,0.49,0.74}
\usepackage[pagebackref,breaklinks,colorlinks,allcolors=cvprblue]{hyperref}

\def\paperID{*****} 
\def\confName{CVPR}
\def\confYear{2026}

\title{{\ourmethod}: Instant 3D Scene Editing from Sparse Unposed Images}

\author{
Jiageng Liu$^{*}$\\
University of Massachusetts Amherst\\
{\tt\small jiagengliu@umass.edu}
\and
Weijie Lyu$^{*}$\\
University of California, Merced \\
{\tt\small wlyu3@ucmerced.edu}
\and
Xueting Li\\
NVIDIA\\
{\tt\small xuetingl@nvidia.com}
\and
Yejie Guo\\
Shanghai Jiao Tong University\\
{\tt\small gyj123@sjtu.edu.cn}
\and
Ming-Hsuan Yang\\
University of California, Merced \\
{\tt\small myang37@ucmerced.edu}
}
\newcommand{\ourmethod}{\textit{Edit3r}}
\newcommand{\ourbench}{\textit{DL3DV-Edit-Bench}}

\begin{document}

\twocolumn[{%
\renewcommand\twocolumn[1][]{#1}%
\maketitle
\begin{center}
\vspace{-8mm}
\centering 
\includegraphics[width=0.95\linewidth]{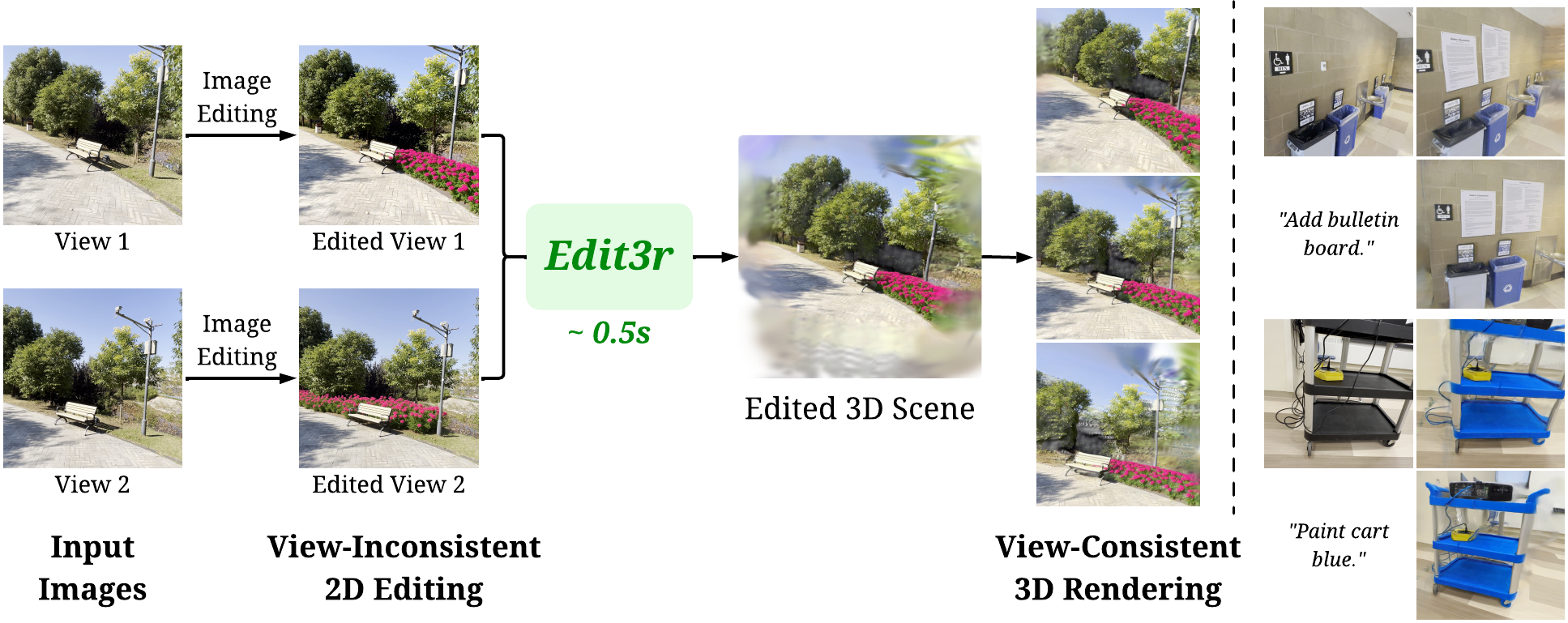}
    \captionsetup{width=\linewidth}
    \captionof{figure}{\textbf{Overview.} {\ourmethod} takes view-inconsistent, 2D edited images as input and generates a view-consistent 3D scene for novel view synthesis in only 0.5 seconds. It supports diverse editing tasks, including insertion, removal, color changes, and weather changes, \etc.}
    \label{fig:teaser}
    \vspace{4mm}
\end{center}
}]
\begingroup
\renewcommand{\thefootnote}{\fnsymbol{footnote}}
\footnotetext[1]{These authors contributed equally.}
\endgroup

\begin{abstract}
We present \textbf{\ourmethod}, a feed-forward framework that reconstructs and edits 3D scenes in a single pass from unposed, view-inconsistent, instruction-edited images.
Unlike prior methods requiring per-scene optimization, {\ourmethod} directly predicts instruction-aligned 3D edits, enabling fast and photorealistic rendering without optimization or pose estimation.
A key challenge in training such a model lies in the absence of multi-view consistent edited images for supervision. We address this with (i) a SAM2-based recoloring strategy that generates reliable, cross-view-consistent supervision, and (ii) an asymmetric input strategy that pairs a recolored reference view with raw auxiliary views, encouraging the network to fuse and align disparate observations.
At inference, our model effectively handles images edited by 2D methods such as InstructPix2Pix, despite not being exposed to such edits during training.
For large-scale quantitative evaluation, we introduce \textbf{{\ourbench}}, a benchmark built on the DL3DV test split, featuring 20 diverse scenes, 4 edit types and 100 edits in total.
Comprehensive quantitative and qualitative results show that {\ourmethod} achieves superior semantic alignment and enhanced 3D consistency compared to recent baselines, while operating at significantly higher inference speed, making it promising for real-time 3D editing applications.
\textbf{Project page:} \url{https://edit3r.github.io/edit3r/}.
\end{abstract}

\vspace{-2mm}
\section{Introduction}
\label{sec:intro}
3D scene editing is a long-standing problem in computer vision. By offering intuitive control over content creation, it plays a crucial role in diverse applications such as visual effects~\cite{zhang2025advancing3dgaussiansplatting}, augmented and virtual reality (AR/VR)~\cite{madhavaram2024trainingfreeapproach3d}, game design~\cite{xu2024sketch2sceneautomaticgenerationinteractive}, and digital twins~\cite{melnik2025digitaltwingenerationvisual}.
Recent advancements in 3D scene representation and reconstruction~\cite{nerf, 3dgs} have sparked a surge of interest in editing 3D scenes using text prompts. Existing methods~\cite{in2n, gaussctrl, lee2025editsplatmultiviewfusionattentionguided} typically combine neural reconstruction with text-driven image editing in a \textit{reconstruct-edit-refit} pipeline. Given a collection of posed images and a textual instruction, these approaches first \textit{reconstruct} the scene using a neural 3D representation, then apply 2D generative \textit{editing} to the rendered images, and finally \textit{refit} (or re-optimize) the 3D representation to incorporate the edits.
While this pipeline achieves impressive visual quality and alignment with instruction prompts, it is limited by slow per-scene optimization and retraining, which limits practical usability. Moreover, these methods often introduce multi-view inconsistencies, revealing a substantial gap between current 3D editing frameworks and the need for fast, interactive scene manipulation in real-world applications.

Conversely, the rapid development of Large Reconstruction Models (LRMs)~\cite{lrm, gslrm, noposplat, mvsnerf, pixelsplat, dust3r} offers a more efficient paradigm for 3D scene reconstruction by amortizing computation across large datasets and removing per-scene optimization. Notably, Gaussian-based LRMs enable rapid, photorealistic rendering suited for real-time applications, thanks to the Gaussian splatting technique.

Motivated by these advancements, we propose {\ourmethod}, a feed-forward approach that reconstructs and edits a 3D scene in a single pass.
Beginning with a pair of unposed views, we edit them using a 2D image editor. These potentially inconsistent edited views serve as inputs for {\ourmethod}, which directly predicts 3D Gaussian splats aligned with the edited views while eliminates the inconsistency, seamlessly unifying reconstruction and editing without the need for per-scene optimization.

Training such a feed-forward reconstruction-editing model introduces two practical challenges. First, ground-truth edited images for supervision are unavailable. To address this, we leverage SAM2-based recoloring~\cite{ravi2024sam} for cross-view-consistent supervision, providing reliable labels and masks. We find that our model, though trained only on a recoloring task, it effectively handles images edited by 2D diffusion models at inference. Second, edited inputs can remain semantically and visually inconsistent across views. To tackle this, we design an asymmetric input scheme that pairs recolored reference views with other views in their original appearance, encouraging the network to fuse and align edited and unedited observations.

Moreover, the lack of established benchmarks limits fair and scalable evaluation in 3D scene editing. We introduce {\ourbench}, a standardized benchmark built on the DL3DV~\cite{ling2023dl3dv10klargescalescenedataset} test split to systematically assess multi-view consistency and inference efficiency, setting a foundation for reproducible and rigorous comparisons.
Our contributions are summarized as follows:
\begin{itemize}
\item We introduce {\ourmethod}, a feed-forward framework that simultaneously reconstructs 3D scenes and generates edited Gaussian splats aligned to a text prompt.
\item We develop a SAM2-based recoloring technique for view-consistent supervision during training and introduce an asymmetric input scheme to bolster robustness against cross-view inconsistencies.
\item We present {\ourbench}, a scene-level, multi-view 3D editing benchmark built on the DL3DV test split.
\item Extensive quantitative and qualitative experiments show {\ourmethod} is significantly faster than previous approaches, achieving higher visual quality and semantic consistency while demonstrating practical real-world potential.
\end{itemize}

\section{Related Work}
\label{sec:related_work}
\subsection{2D Image and Video Editing}
Recent diffusion-based editors enable high-quality, instruction-conditioned edits through techniques such as noise-to-image guidance~\cite{meng2022sdeditguidedimagesynthesis}, cross-attention manipulation~\cite{hertz2022prompttopromptimageeditingcross}, instruction tuning~\cite{ip2p}, and structure-conditional control~\cite{controlnet}. While effective for single images, naively applying them per frame leads to flicker and identity drift.

To address temporal coherence, video extensions employ attention/token propagation~\cite{geyer2023tokenflowconsistentdiffusionfeatures}, edit-specific attention alignment~\cite{qi2023fatezerofusingattentionszeroshot,liu2023videop2pvideoeditingcrossattention}, optical-flow guidance~\cite{yang2023rerendervideozeroshottextguided}, and implicit canonicalization~\cite{ouyang2024codefcontentdeformationfields}. Despite strong short-range coherence, these approaches remain fundamentally 2D and lack explicit 3D scene reasoning, leading to view-inconsistent artifacts when lifted to 3D supervision.

\subsection{3D Scene Editing}
Neural reconstruction methods such as NeRF~\cite{nerf} and 3D Gaussian Splatting~\cite{3dgs} enable photorealistic rendering, motivating integration with semantic editing. Existing approaches typically follow a reconstruct-edit-refit pipeline. Instruct-NeRF2NeRF~\cite{in2n} reconstructs scenes with NeRF, edits multi-view renderings using InstructPix2Pix~\cite{ip2p}, and re-optimizes the scene with edited images. GaussianCtrl~\cite{gaussctrl} employs ControlNet~\cite{controlnet} with depth conditioning and attention-based alignment for view consistency, while GaussianEditor~\cite{gaussianeditor} introduces Gaussian semantic tracing for precise editing control. Though achieving impressive quality, these methods require slow per-scene optimization and are prone to multi-view inconsistencies.

\subsection{Generalizable Feed-forward Reconstructors}
Large reconstruction models (LRMs)~\cite{mvsnerf, pixelsplat, mvsplat, noposplat, gslrm, lgm, splatter_image, lsm} have revolutionized 3D reconstruction by learning generalizable priors from large-scale datasets like Objaverse~\cite{objaverse}. Leveraging transformer architectures~\cite{transformer, vit}, these models enable rapid inference of NeRF or Gaussian representations from sparse views without per-scene optimization. Methods such as PixelSplat~\cite{pixelsplat} utilize epipolar geometry, while many other methods~\cite{mvsnerf, murf, mvsplat, mvsplat360} construct cost volumes for multi-view aggregation. Despite their efficiency in reconstruction, the application of feed-forward reconstructors to 3D scene editing remains largely unexplored, which our work addresses.
\begin{figure*}[t]  
    \centering
    \includegraphics[width=\textwidth]{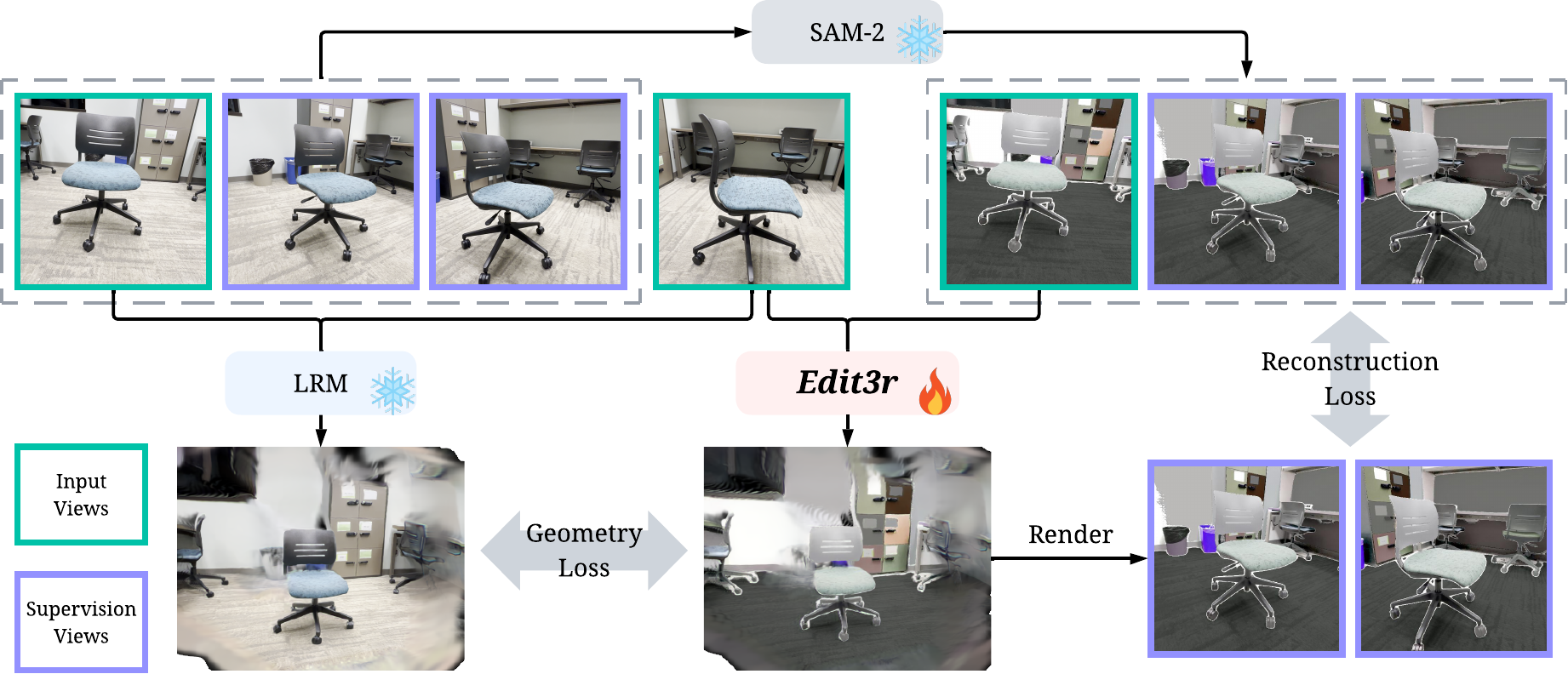}
    \caption{\textbf{Training pipeline of \ourmethod.}
    Green-framed images are input views and purple-framed images are supervision views (ground-truth and rendered results).
    We first apply SAM2-based recoloring to both inputs and supervision views, then feed an asymmetric pair (one recolored view and one original view) into {\ourmethod} to predict a 3D Gaussian scene, which is rendered and supervised with reconstruction losses against the recolored supervision views (right).
    In parallel, a frozen LRM reconstructs the scene from the original inputs and provides 3D supervision via geometry losses that regularize {\ourmethod}’s Gaussian predictions in 3D space.}
    \label{fig:pipeline}
    \vspace{-2mm}
\end{figure*}
\section{Method}
We propose a novel feed-forward framework for 3D scene editing. Given multi-view images and a text prompt, our method first applies 2D edits~\cite{ip2p, labs2025flux1kontextflowmatching, gpt4o} to the input images, and then reconstructs a 3D scene that is aligned with the text prompt, even when the edited images introduce multi-view inconsistencies. In contrast to optimization-based approaches~\cite{in2n, gaussctrl, gaussianeditor}, our pipeline enables real-time, instruction-guided editing by leveraging the efficient reconstruction capabilities of LRMs. Furthermore, {\ourmethod} is highly modular and can seamlessly integrate any 2D image editing technique, such as Instruct-Pix2Pix~\cite{ip2p} and FLUX~\cite{labs2025flux1kontextflowmatching}, thereby substantially enhancing the diversity and flexibility of the generated results.
\subsection{Preliminaries}
Given a sequence of unposed images with corresponding camera intrinsics $\{(I_v, k_v)\}_{v=0}^{V-1}$ and an editing text prompt $T$, where $V$ is the number of input views, our goal is to reconstruct a geometrically consistent 3D scene $S_T$ that is semantically aligned with $T$.

During training, our pipeline takes one recolored image $(I'_0, k_0)$ and another original image $(I_1, k_1)$ from the unposed sequence as inputs to a feed-forward reconstruction model $f_{\theta}$ with learnable parameters $\theta$.
Instead of explicitly estimating camera poses or meshes, we approach reconstruction by lifting the unposed images into a canonical 3D Gaussian scene. Given views with unknown extrinsics, the network predicts a set of anisotropic Gaussian primitives in a fixed world frame. Each primitive encodes both geometry (location and shape) and appearance attributes (radiance and opacity).
Formally, we learn a function that maps image evidence to this representation:
\begin{equation*}
\mathcal{G} \;=\; f_{\theta}\!\left(\{(I'_0, k_0),(I_1, k_1)\}\right)
\end{equation*}
where $\mathcal{G}=\{(\mu_j,\Sigma_j,c_j,\alpha_j)\}_{j=0}^{V-1}$ denotes the predicted set of 3D Gaussians. $\mu\!\in\!\mathbb{R}^3$ is the center, $\Sigma\!\in\!\mathbb{R}^{3\times3}$ is the covariance, 
$c$ is a vector of spherical-harmonic color coefficients, and $\alpha$ is opacity.

\vspace{1mm}
\noindent\textbf{Pose-Free 3D Scene Reconstruction.}
Our approach adopts the pose-free 3D reconstruction framework of NoPoSplat~\cite{noposplat}. Given sparse unposed multi-view images, we embed the camera intrinsics for each view using a small MLP $\phi:\mathbb{R}^d \to \mathbb{R}^D$ concatenate these embeddings with the image token sequences to obtain the input tokens $z_v$:
\begin{equation*}
    z_v = \big[\,\text{img\_tokens}(I_v) \oplus \phi(k_v)\,\big],
\end{equation*}
A vision transformer (ViT)~\cite{vit} encoder with shared weights processes each view independently, taking $z_v$ as input and producing per-view features $f_v$ in a unified feature space. The set of features $\{f_v\}_{v=0}^{V-1}$ is then fused using a ViT decoder that leverages both self-attention and cross-view attention, enabling the model to resolve occlusions and appearance variations between views. This decoder outputs fused features $\{f_{fused}\}_{v=0}^{V-1}$, which are subsequently passed to two lightweight Gaussian heads.
Each Gaussian head contains two DPT-based~\cite{dust3r} predictors: one predicts the 3D centers of the Gaussians using only transformer features to ensure geometric stability; the other incorporates both the transformer features and RGB image shortcuts to estimate additional attributes, such as opacity, covariance, and low-order spherical harmonics for view-dependent color. For every input, the model produces a dense set of Gaussian primitives, which are concatenated in a canonical frame and rendered by standard 3D Gaussian splatting. The model is trained using photometric losses, eliminating the need for explicit pose input or pose-based warping and resulting in a pose-free, feed-forward pipeline.

\begin{figure}[t]  
    \centering
    \includegraphics[width=\linewidth]{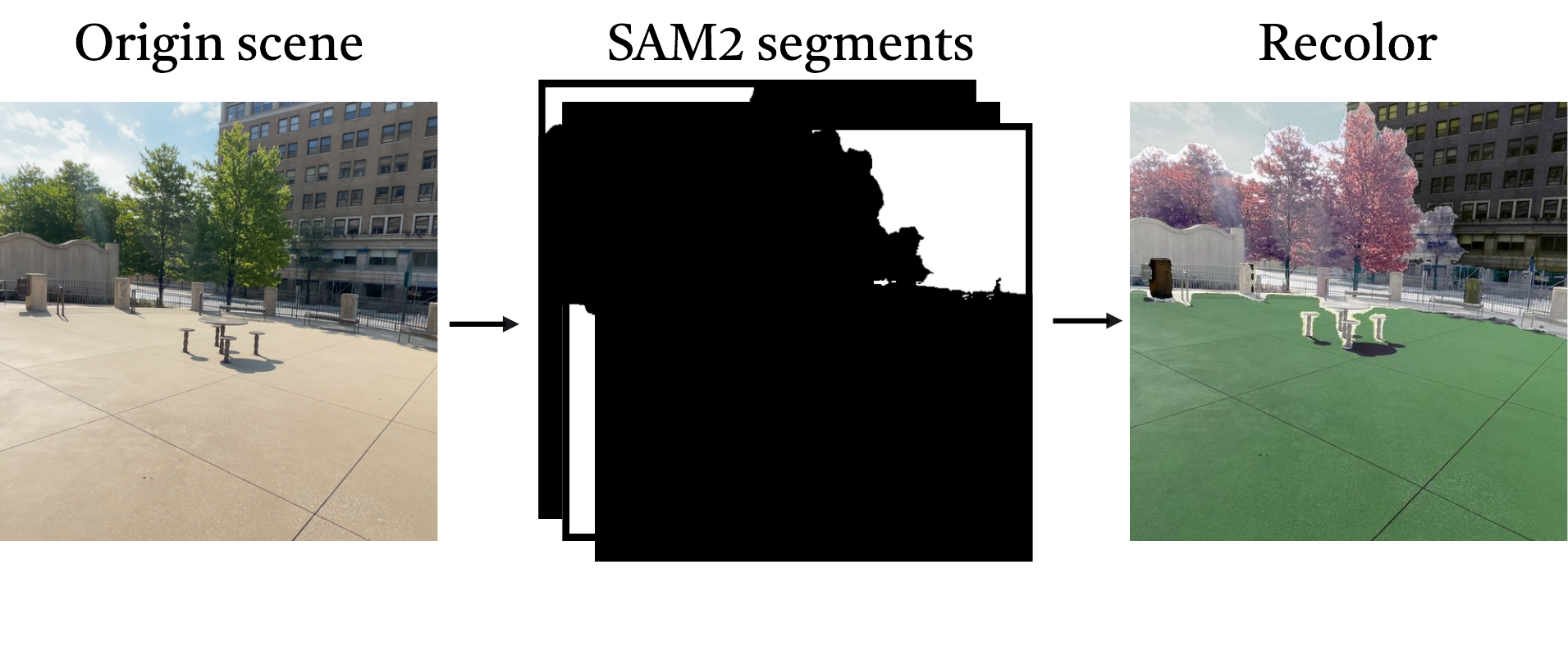}  
    \vspace{-2mm}
    \caption{Example of SAM2-based recoloring process.}
    \label{fig:SAM2-recolor}
    \vspace{-4mm}
\end{figure}

\subsection{SAM2-Based Recoloring} 
A key challenge in training our feed-forward 3D scene editing framework is the lack of multi-view consistent images paired with realistic edited inputs for supervision. To overcome this, we propose to approximate the editing task with recoloring. Specifically, we build upon SAM2~\cite{ravi2024sam} and design an object-aware segmentation and recoloring pipeline: the recoloring branch generates multi-view consistent supervision targets, while SAM2’s masks enable object-level appearance augmentations. These object-level augmentations play a crucial role in bridging the gap between our recolored training data and the edited images encountered at inference, ultimately enhancing the model’s generalization and performance. The overall data processing pipeline is illustrated in Fig.~\ref{fig:SAM2-recolor}.

\vspace{1mm}
\noindent\textbf{Object-Level Multi-Frame Segmentation via SAM2.} Given multi-view video frames $\{I_v\}_{v=0}^{V-1}$, we apply the automatic mask generator (AMG) from SAM2 on the first frame $I_0$ to automatically identify all candidate objects, without the need for manual prompts. Each AMG proposal $m$ provides a soft mask, area, stability score, predicted IoU, and a bounding box $b=(x, y, w, h)$, enabling efficient object selection and subsequent recoloring across views. 

We keep a proposal if it passes a deterministic multi-criterion filter:
(1) area $\ge A_{\min}$; 
(2) stability $\ge s_{\min}$; 
(3) predicted IoU $\ge q_{\min}$; 
(4) aspect ratio $w/h$ within $[r_{\min}, r_{\max}]$; 
(5) the box has at least $m_{\mathrm{edge}}$ pixels of margin to the image boundary. 
These parameter details are provided in the appendix. All retained instances are assigned persistent IDs and used to prompt the SAM2 video segmentation predictor (VSP) at $t{=}0$.

Next, we apply the VSP to the video sequence. For each frame $I_{v}$, the model identifies the set of active object IDs $O_v$ and corresponding per-object mask logits $Z_v^{(r)}$ which indicate foreground probabilities. Binary masks are obtained by thresholding the logits at zero, with pixels having logits greater than zero designated as foreground.
To mitigate identity drift, we only process a frame $v$ if its object-ID set $O_v$ sufficiently overlaps with that of the most recently processed frame $v^-$. Formally, we define the overlap ratio as $\frac{|O_v \cap O_{v^-}|}{\max(|O_v|, |O_{v^-}|)}$. If this ratio falls below 0.5, frame $I_v$ is skipped and excluded from all subsequent stages.
For all remaining (non-skipped) frames $v$ in the video, this stage produces per-object binary masks $\{M_v^{(r)}\}_{r \in O_v}$, which guide the subsequent region-wise recoloring process.

\vspace{1mm}
\noindent\textbf{Region-Wise Recoloring With Cross-View Consistency.}
We perform region-aware color augmentation by sequentially applying a composite color transform $\mathcal{C}_{\Theta_r}$ to each region. This transform consists of ColorJitter, gamma correction, PCA-based lighting offset, a fixed RGB-channel permutation, and optional grayscale conversion. The augmentation parameters $\Theta_r$ are sampled once per region and then reused across all frames and views to ensure consistent appearance.
When regions overlap, we re-normalize the per-pixel soft masks so that the mask weights sum to one over all overlapping regions (with a small $\varepsilon$ added to the denominator for numerical stability). The recolored image is then synthesized by soft blending:
\begin{equation*}
I_t' \;=\; 
\sum_{r\in\mathcal{R}} \hat{\alpha}_t^{(r)} \odot \mathcal{C}_{\Theta_r}(I_t)
\;+\;
\Big(1 - \sum_{r\in\mathcal{R}} \hat{\alpha}_t^{(r)}\Big) \odot I_t
\end{equation*}
Here, \(\odot\) denotes element-wise multiplication, and \(\hat{\alpha}_t^{(r)}\) is the per-pixel \emph{renormalized} mask weight for region \(r\); the remaining weight $1 - \sum{r\in\mathcal{R}} \hat{\alpha}_t^{(r)}$ at each pixel is assigned to the original image.
With a single region, this reduces to standard alpha blending between the transformed and original image.

\subsection{Loss Functions}
We train the model using both 2D supervision signals and 3D geometric constraints.  
For each view $v$, given the rendered prediction $\hat I_v$ produced by our model and the corresponding ground-truth image $I_v$, 
we optimize the model parameters by minimizing a weighted sum of three complementary loss terms across all views:
\begin{equation*}
\begin{split}
\label{eq:edit-loss}
\min_{\theta}\; \sum_{v=0}^{V-1}\Big[
&\mathcal{L}_{\text{CLIP}}(\hat I_v, I_v)
+\mathcal{L}_{\text{LPIPS}}(\hat I_v, I_v)
+\mathcal{L}_{\text{MSE}}(\hat I_v, I_v)
\Big] \\
&+\, \mathcal{L}_{\text{center}}
+\, \mathcal{L}_{\text{geom}}.
\end{split}
\end{equation*}
where the terms are weighted by their coefficients respectively
%
The loss functions are defined as follows:  
a) a CLIP-based image-image loss $\mathcal{L}_{\text{CLIP}}(\hat I_v, I_v)$ that aligns the predictions with the targets in a semantic embedding space, providing a robust global supervision signal;  
b) a VGG-based perceptual loss $\mathcal{L}_{\text{LPIPS}}(\hat I_v, I_v)$ that enhances mid- and high-frequency details such as edges and textures, improving perceptual fidelity beyond pixel-level matching; and  
c) a low-frequency MSE loss $\mathcal{L}_{\text{MSE}}(\hat I_v, I_v)$ that enforces consistency in color, exposure, and illumination while remaining tolerant to small reprojection errors.

Beyond the 2D rendering losses, we introduce 3D regularization on the predicted Gaussian centers to stabilize the scene structure during editing.  
Because the editing applied to input images may distort depth cues and cause geometric inconsistencies across views, we leverage the pretrained LRM (NoPoSplat~\cite{noposplat}) to extract reference Gaussian centers $\mathcal{G}_{\text{ref}}$ from the unedited images.  
During training, the edited Gaussians $\mathcal{G}_{\text{edit}}$ are constrained to align with these reference centers using two complementary regularization terms.
First, a center-matching loss encourages each predicted Gaussian to remain close to its corresponding reference location, formulated as a Huber loss~\cite{huber1964robust} between the predicted and reference 3D centers:
\begin{equation*}
    \mathcal{L}_{\text{center}} = \mathrm{SmoothL1}(\hat{\boldsymbol{\mu}}, \boldsymbol{\mu}_{\text{ref}}).
\end{equation*}
This term preserves geometric fidelity to the original scene while permitting local deformations consistent with the applied edits.
Second, a multi-view consistency loss enforces structural coherence across different edited views by minimizing the pairwise Chamfer-$L_1$ distance between randomly sampled Gaussian centers from each view:
\begin{equation*}
\mathcal{L}_{\text{geom}} = \frac{1}{V(V-1)} 
\sum_{i<j}
\mathrm{Chamfer}_{L_1}(\hat{\boldsymbol{\mu}}_i,\hat{\boldsymbol{\mu}}_j)
\end{equation*}
This encourages a consistent 3D configuration across views, reducing depth-layer separation and suppressing “floating” artifacts caused by view-dependent misalignment.
Together, $\mathcal{L}_{\text{center}}$ anchors the edited scene to the base geometry learned by the pretrained reconstruction model, while $\mathcal{L}_{\text{geom}}$ encourages smooth geometric alignment across views.  
These 3D regularization terms complement the 2D perceptual and semantic losses by enforcing structural stability in the canonical Gaussian space, ensuring that scene edits remain spatially coherent and physically plausible.

\begin{figure}[t]
    \centering
    \includegraphics[width=\linewidth]{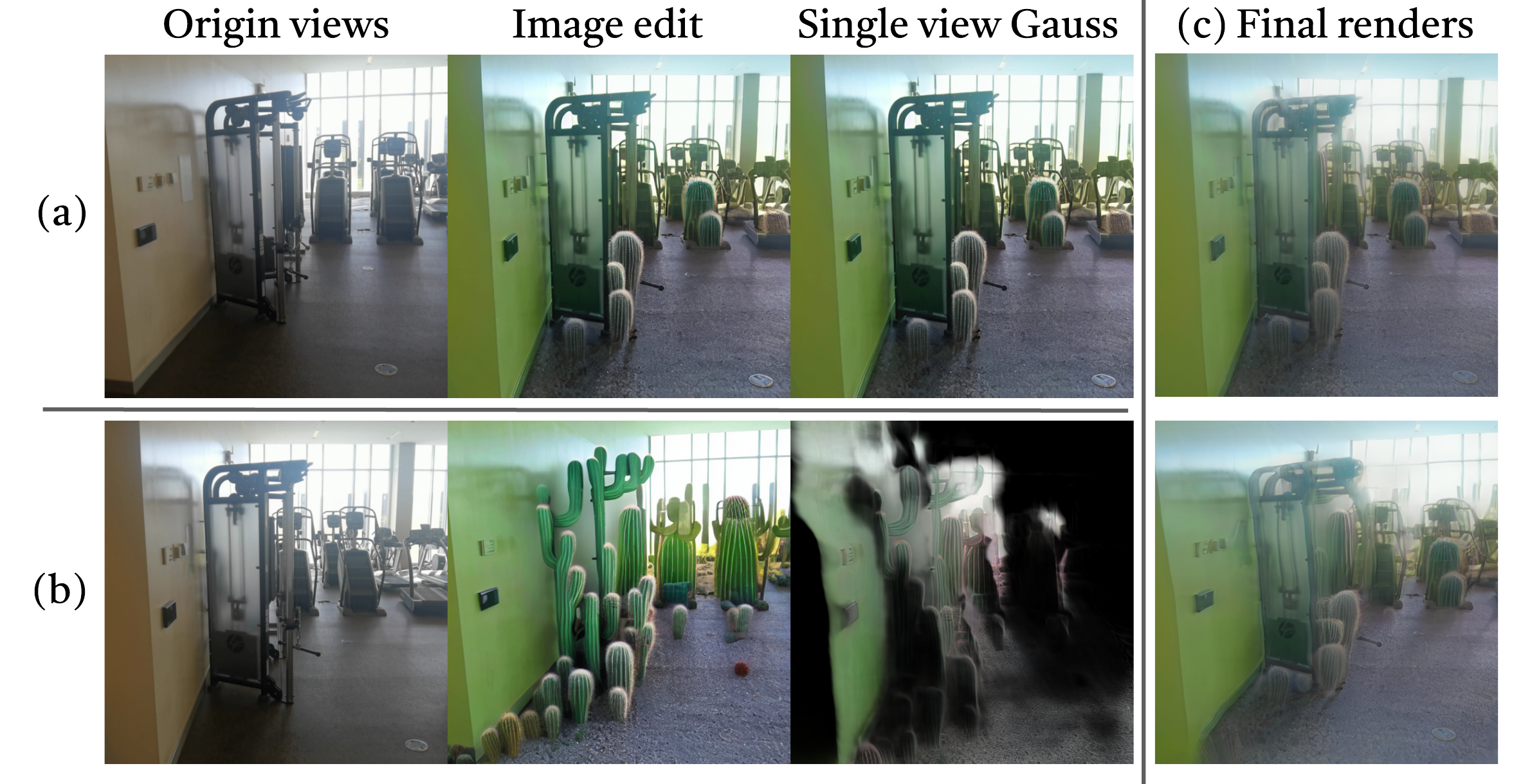}  
    \caption{\textbf{Example of inference-time editing.}
    (a) First input view, its edited result, and the corresponding single-view Gaussian rendering.
    (b) Second input view, its edited result, and the corresponding single-view Gaussian rendering.
    (c) Final rendering obtained by combining Gaussians from both views. \textit{Prompt: Add a cactus garden.}}
    \label{fig:edit_example}
    \vspace{-2mm}
\end{figure}
\subsection{Inference}
Although our model is trained solely on recolored images, it generalizes effectively to images edited by various 2D editing methods~\cite{ip2p, labs2025flux1kontextflowmatching}.  
Unlike optimization-based pipelines that require iterative test-time fitting, inference in our system is fully feed-forward.  
Since the 2D editing stage operates independently from both the feed-forward reconstruction and the training procedure, our approach is highly flexible and can accommodate diverse image editing techniques~\cite{gpt4o, labs2025flux1kontextflowmatching, geminiteam2025geminifamilyhighlycapable}.  
For experiments and evaluations, we primarily employ commonly used IP2P~\cite{ip2p} as the 2D editor, while additional editors are included in ablation studies.  

As illustrated in Fig.~\ref{fig:teaser}, given a multi-view sequence \(\{I_v\}_{v=0}^{V-1}\) and a text instruction \(\mathcal{E}\), we first generate per-frame edited images \(\{I_v^\star\}_{v=0}^{V-1}\) with a 2D image editor.  
This differs from the training setup, where only the first frame is edited; at inference, all frames are edited to maximize cross-view evidence.  
We then perform a single feed-forward pass of {\ourmethod} to produce the edited 3D Gaussians and render novel views.
Fig.~\ref{fig:edit_example} provides a more concrete example: we visualize the Gaussians generated from the two input views separately in (a) and (b). 
Due to the inconsistency introduced by the 2D image editor, pixels in the second view can conflict with those in the first view; in these regions, the Gaussians from the second view automatically reduce their opacity to mitigate the conflict. 
Panel (c) shows the final rendering obtained by combining the Gaussians from both views, demonstrating view-consistent results.

\subsection{DL3DV-Edit-Bench}
\vspace{1mm}
\noindent\textbf{Motivation and Scope.}
Most existing 3D editing evaluations~\cite{in2n} either focus on single-object or toy scenes rather than full scene editing (\eg, limited context and occlusions), or rely on ad-hoc, privately curated data without a standardized protocol. This hampers fair comparison: methods differ in scene diversity, edit categories, and evaluation criteria. 
We therefore build \textbf{{\ourbench}} on top of the DL3DV~\cite{ling2023dl3dv10klargescalescenedataset} test split, covering diverse indoor/outdoor real scenes and four text-driven edit types: \emph{Add}, \emph{Remove}, \emph{Modify} (attribute/material/appearance), and \emph{Global} (style/exposure). The benchmark targets \emph{scene-level} editing under multi-view inputs and measures both edit effectiveness and cross-view consistency in realistic settings.

\vspace{1mm}
\noindent\textbf{Generation Pipeline.}
We start from the DL3DV test set and sample $20$ real-world scenes (indoor / outdoor). We first run Grounding-DINO~\cite{liu2024groundingdinomarryingdino} on all views to obtain object-level labels and region proposals. The per-scene image set and extracted labels are then fed to a large language model to synthesize candidate prompts for the four edit categories (Add / Remove / Modify / Global). 
We manually vet and keep $5$ valid prompts per scene, resulting in $100$ total edit instances. To ensure multi-view consistency during editing, we fix the 2D image editor, \emph{share} the same random seed and textual prompt across all views of a scene, and confine local edits with masks derived from the proposals. All editor hyperparameters (CFG, steps, strength schedule) are documented and kept constant unless otherwise stated.

\section{Experiments}

\subsection{Implementation Details}
Our main experiments are conducted on the {\ourbench} that we curate to serve as our primary evaluation benchmark. 
For the 2D image-editing module, we use InstructPix2Pix~\cite{ip2p} as the default editor to ensure comparability, and we evaluate more editors in ablations. All experiments are run on a single NVIDIA RTX 6000 GPU.

\subsection{Baselines}
We compare against three methods: (1) EditSplat~\cite{lee2025editsplatmultiviewfusionattentionguided} (optimization-based) with its original 2D editor, InstructPix2Pix, kept unchanged to avoid re-tuning and ensure fairness; (2) GaussCtrl~\cite{gaussctrl} (optimization-based) likewise retaining the authors’ ControlNet module. Both optimization-based baselines are compute-intensive and, in addition to the text prompt, require a pre-reconstructed DL3DV scene, which we supply from standardized baseline reconstructions for parity. (3) NoPoSplat~\cite{noposplat} (feedforward-based), which runs the released checkpoint and consumes only the multi-view inputs for reconstruction, serving as a lower bound on cross-view edit consistency and isolating the contribution of our cross-view fusion in the feed-forward design.

\begin{figure*}[t]
    \centering
    \includegraphics[width=0.93\textwidth]{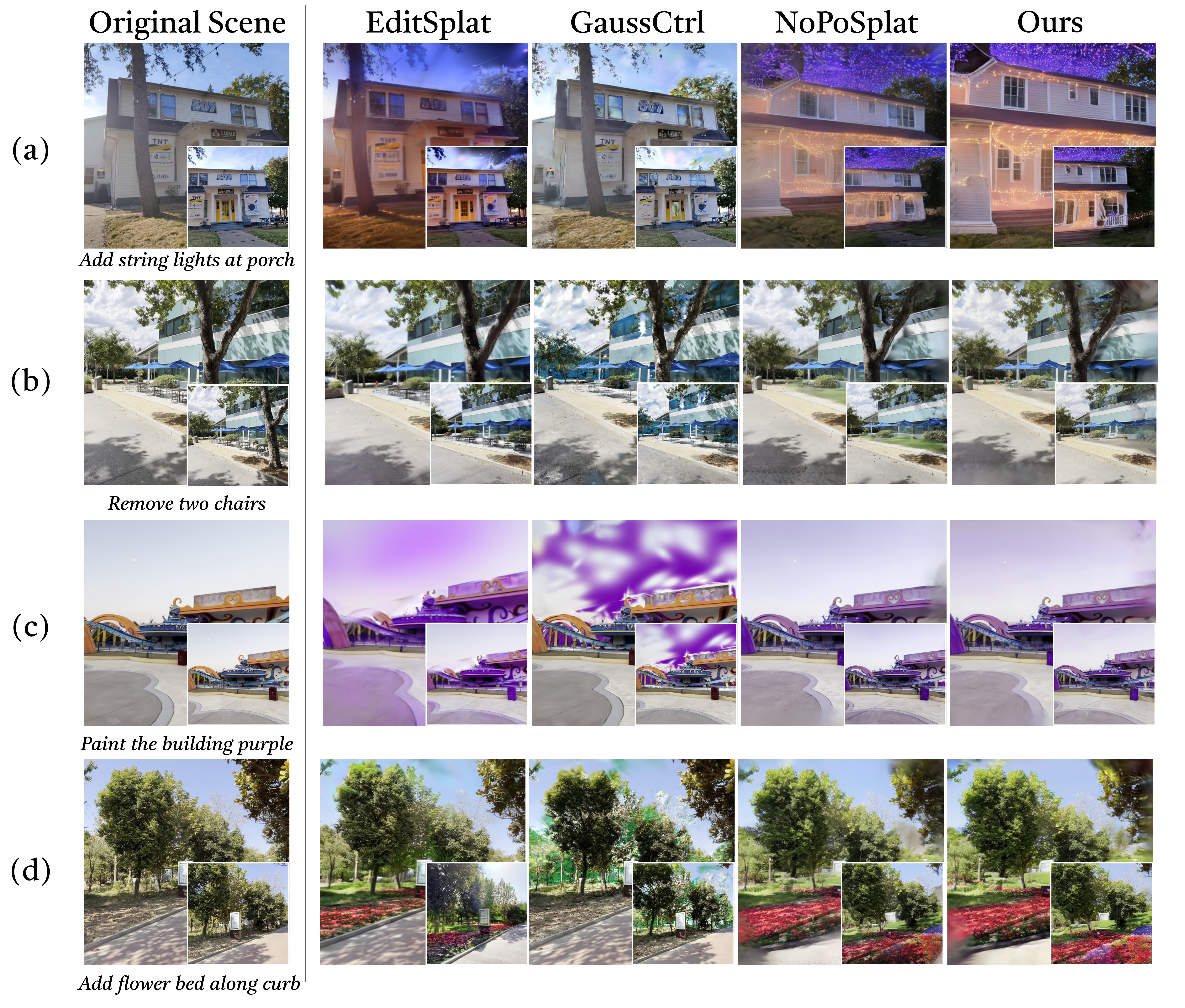}
    \caption{\textbf{Qualitative comparison} among four methods. The first row shows the original scene and the editing prompt; the remaining four rows show the results, each paired with an additional view.}
    \label{fig:qualitative-examples}
\end{figure*}

\subsection{Main Experiments} 
\vspace{1mm}
\noindent\textbf{Evaluation Metrics.} Similar to prior works~\cite{lee2025editsplatmultiviewfusionattentionguided, gaussctrl, in2n, you2025instainpaintinstant3dsceneinpainting}, we evaluate effectiveness from two angles: (1) whether the method performs the intended scene edits; (2) whether the 3D reconstruction is plausible across viewpoints. For (1), acknowledging the subjectivity of 3D scene generation, we adopt CLIP~\cite{radford2021learningtransferablevisualmodels} image-text similarity, which scores the absolute alignment between the edited render and the target description. For (2), we report C-FID~\cite{heusel2018ganstrainedtimescaleupdate}, which measures distribution-level realism by computing the Fréchet distance between feature distributions of edited renders and reference views, and C-KID~\cite{bińkowski2021demystifyingmmdgans}, which measures the kernel MMD between the same feature distributions and is more reliable with moderate sample sizes. Here, the prefix "C-" indicates that both metrics are computed in a scene-conditioned manner. Together they quantify overall realism/naturalness across views.

\vspace{1mm}
\noindent\textbf{Quantitative Comparison.} As summarized in Tab.~\ref{tab:main}, {\ourmethod} achieves the best quality-speed trade-off.
In terms of runtime, feed-forward methods are dramatically faster than optimization-based ones; 
{\ourmethod} processes a view in 0.51\,s (vs.\ 325.53\text{–}584.46\,s for optimization-based baselines). 
{\ourmethod} also attains the highest CLIP$_{t2i}$(text prompts to rendered images), indicating stronger instruction following and more effective scene edits. 
By design, GaussianCtrl applies only minor modifications, which keeps outputs close to the original and thus boosts C-FID and C-KID, but it underperforms on text alignment due to limited edit strength. 
EditSplat makes moderate edits and is relatively stable, yet remains conservative overall. 
NoPoSplat focuses on reconstruction rather than explicit editing; when input views are inconsistent, it tends to average conflicting evidence, producing blurred renders and uniformly lower scores (CLIP$_{t2i}$, C-FID and C-KID). 
Overall, {\ourmethod} best balances instruction following with multi-view realism, improving C-FID and C-KID over other feed-forward baselines while preserving semantics and structure nearly as well as the most conservative optimizer.

\vspace{1mm}
\noindent\textbf{Qualitative Comparison.}
Fig.~\ref{fig:qualitative-examples} presents four examples (a–d) of scene editing by EditSplat~\cite{lee2025editsplatmultiviewfusionattentionguided}, GaussCtrl~\cite{gaussctrl}, NoPoSplat~\cite{noposplat}, and {\ourmethod}. The small inset at the bottom-right of each panel shows the result from another viewpoint of the same scene. EditSplat achieves partially successful edits in (a, c, d), but it also alters regions that should remain unchanged, such as the sky and ground in (c). GaussCtrl is almost entirely unsuccessful across (a–d), producing chaotic or ineffective edits when applied to scenes that are unseen during its training. NoPoSplat generally generates plausible results, but its quality is highly sensitive to the inconsistency between input views: when cross-view inconsistency is large, it suffers from blur in (a,d), whereas for more consistent inputs in (b,c), the reconstruction quality is much better. In contrast, our method delivers stable, high-quality edits across all four scenes, maintaining consistency across views while preserving the original scene structure.

\begin{table}[t]
\centering
\scriptsize
\caption{\textbf{Quantitative comparison} on optimization-based and feed-forward methods.}
\label{tab:main}
\setlength{\tabcolsep}{6pt}
\renewcommand{\arraystretch}{1.1}
\begin{tabular}{l@{\hskip 1.0mm}l@{\hskip 1.0mm}c@{\hskip 1.0mm}c@{\hskip 1.0mm}c@{\hskip 1.0mm}c}
\toprule
\textbf{Category} & \textbf{Method} &
\textbf{Time (s)}\,(\(\downarrow\)) &
\text{CLIP}$_{t2i}$(\(\uparrow\)) &
\textbf{C-FID}\,(\(\downarrow\)) &
\textbf{C-KID}\,(\(\downarrow\)) \\
\midrule
\multirow{2}{*}{Optimization}
& GaussCtrl~\cite{gaussctrl} & 325.53 & 0.227 & 135.0 & 0.091 \\
& Editsplat~\cite{lee2025editsplatmultiviewfusionattentionguided} & 584.46 & 0.241 & 174.1 & 0.122 \\
\midrule
\multirow{2}{*}{Feed-forward}
& NoPoSplat~\cite{noposplat} & 0.61   & 0.253 & 180.6 & 0.125 \\
& {\ourmethod} (Ours)   & 0.51   & 0.266 & 171.3 & 0.116 \\
\bottomrule
\vspace{-4mm}
\end{tabular}%
\end{table}

\begin{table}[t]
\centering
\scriptsize
\caption{\textbf{Ablation quantitative comparison} on different training strategies in our method.}
\label{tab:training_ablation}
\setlength{\tabcolsep}{6pt}
\renewcommand{\arraystretch}{1.1}
\begin{tabular}{l@{\hskip 8.0mm}c@{\hskip 8.0mm}c@{\hskip 8.0mm}c}
\toprule
\textbf{Method} & \textbf{CLIP$_{t2i}$~(\(\uparrow\))} &
\textbf{C-FID~(\(\downarrow\))} & \textbf{C-KID~(\(\downarrow\))} \\
\midrule
{\ourmethod}  & 0.266 & 171.3 & 0.116 \\
+w/o Recolor   & 0.243 & 215.0 & 0.141 \\ 
+w/o 3D Loss   & 0.237 & 278.4 & 0.182 \\ 
+w/o SAM  & 0.248 & 179.6 & 0.127 \\ 
+w/o R-Drop & 0.252 & 183.1 & 0.130 \\ 
\bottomrule
\vspace{-4mm}
\end{tabular}
\end{table}

\subsection{Ablation Study}
We validate our design choices by comparing {\ourmethod} against different training and inference variants. Unless otherwise stated, the default configuration uses SAM2-based recoloring for supervision, the asymmetric input scheme, our full training losses (2D appearance + 3D regularization), randomly dropping the first view's Gaussian with $p{=}0.5$ during training; and InstructPix2Pix as the editing front-end at inference.

\vspace{1mm}
\noindent\textbf{Recolor/Image Editing For Training.}
Directly training on multi-view images produced by a 2D editor inevitably introduces substantial cross-view inconsistencies that act as label noise and hinder convergence. We therefore supervise training with \emph{SAM2-based recoloring}: object masks are extracted and recolored consistently across views, yielding stable, view-aligned targets. As Table \ref{tab:training_ablation} shows, in our ablations, replacing recoloring data with direct multi-view editing degrades multi-view consistency and increases artifacts, confirming the need for view-consistent supervision.

\vspace{1mm}
\noindent\textbf{Training Loss.} 
Because recoloring/editing can shift predicted Gaussian centers, relying solely on 2D losses (reconstruction/style) permits geometry drift: the rendering may look plausible while Gaussian primitives misalign in 3D. We ablate our 3D regularizers, and results are shown in Tab.~\ref{tab:training_ablation} . Removals lead to larger center deviation and diminished cross-view coherence. Using the full loss stabilizes geometry and improves semantic fidelity without sacrificing appearance.

\vspace{1mm}
\noindent\textbf{SAM2 Segmentation For Augmentation.}
To narrow the distribution gap between recolored training images and edited images at inference, we augment training data with SAM2-derived mask jittering (dilation/erosion), palette perturbation, and limited background leakage. Ablating these SAM2-based augmentations by training with plain recoloring only, reduces robustness and increases failure cases under stronger edits, indicating that modest variability during training is beneficial. Results are shown in Tab.~\ref{tab:training_ablation}.

\vspace{1mm}
\noindent\textbf{Random Drop the First View's Gaussian.}
Given the asymmetric inputs, the edited reference view can dominate the learned style. We introduce random drop during training: with probability $0.5$, we discard the Gaussians associated with the first view when forming supervision, forcing the network to propagate the edit semantics from the reference to the auxiliaries. Ablations on the drop probability, results in Tab.~\ref{tab:training_ablation}, show that removing random drop causes overfitting to the reference viewpoint, while excessively high drop rates weaken edit strength.

\begin{figure}[t]  
\scriptsize
    \centering
    \includegraphics[width=\linewidth]{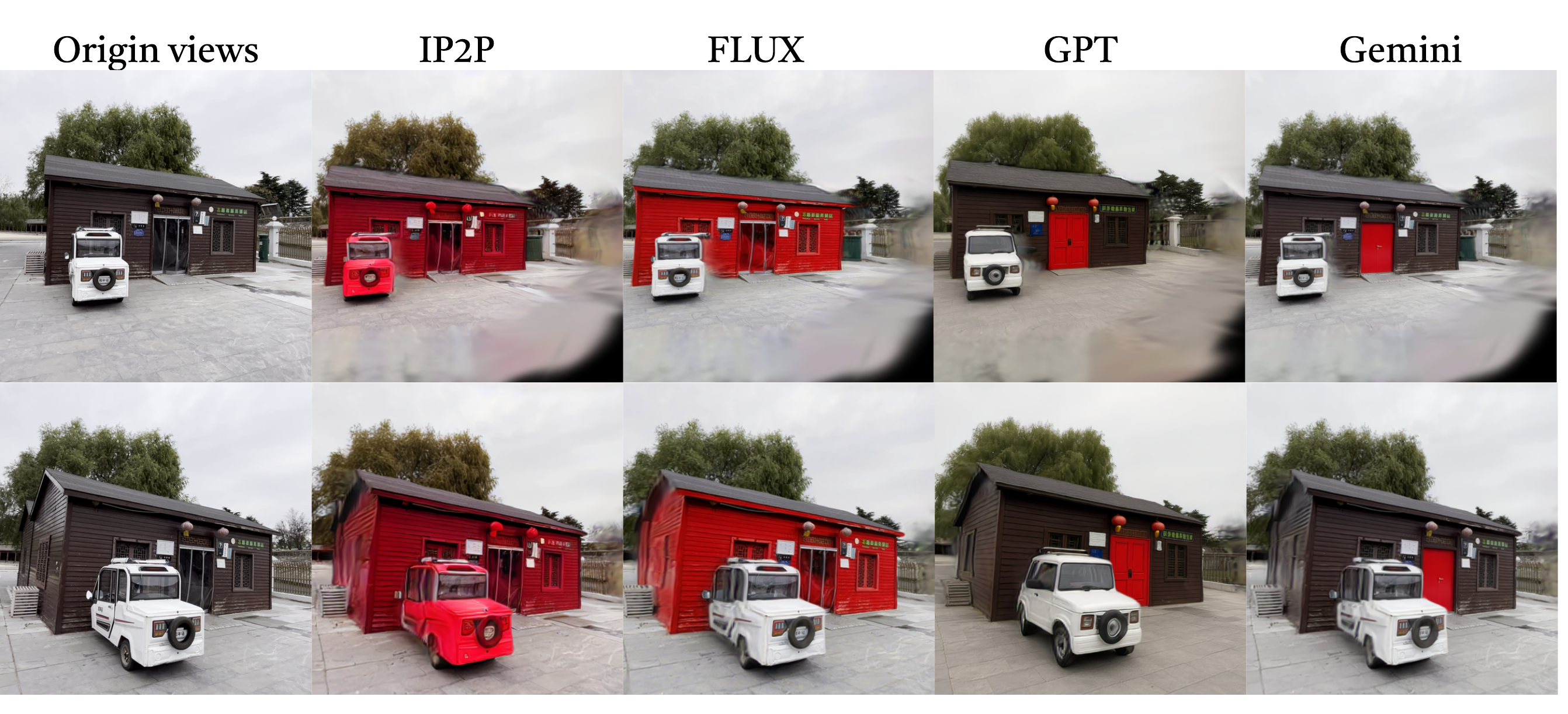}
    \caption{\textbf{Ablation qualitative comparison} among four image editors. \textit{Prompt: Paint the door red.}}
    \label{fig:image-editor}
    \vspace{-4mm}
\end{figure}

\vspace{1mm}
\noindent\textbf{Image Editing For Inference.}
Our pipeline decouples 2D editing from feed-forward reconstruction, allowing different editors at test time. We evaluate IP2P~\cite{ip2p} and alternatives including (1) OpenAI GPT-Image-1~\cite{gpt4o} (GPT) (2) Google Gemini-2.5-Flash-Image~\cite{geminiteam2025geminifamilyhighlycapable} (Gemini) (3) FLUX~\cite{labs2025flux1kontextflowmatching}. 
As shown in Fig.\ref{fig:image-editor} and Tab.\ref{tab:image_editor}, FLUX produces the highest visual edit quality but its edited regions are sometimes mislocalized; GPT and Gemini accurately identify the intended edit regions but tend to modify other parts of the image, while IP2P yields weak perceptual edit quality among the four.
Across editors, {\ourmethod} preserves its advantages in speed and consistency; stronger editors can improve local edit quality, while our model continues to enforce cross-view coherence. This indicates that {\ourmethod} is editor-agnostic and can benefit from future improvements in 2D editing.

\begin{table}[t]
\centering
\scriptsize
\caption{\textbf{Ablation quantitative comparison} on different image editors in our method.}
\vspace{-2mm}
\label{tab:image_editor}
\setlength{\tabcolsep}{6pt}
\renewcommand{\arraystretch}{1.1}

\begin{tabular}{l@{\hskip 8.0mm}c@{\hskip 8.0mm}c@{\hskip 8.0mm}c}
\toprule
\textbf{Method} & \textbf{CLIP$_{t2i}$~(\(\uparrow\))} &
\textbf{C-FID~(\(\downarrow\))} & \textbf{C-KID~(\(\downarrow\))} \\
\midrule
IP2P~\cite{ip2p}  & 0.266 & 171.3 & 0.116 \\
GPT~\cite{gpt4o}    & 0.261 & 166.2 & 0.102 \\ Gemini~\cite{geminiteam2025geminifamilyhighlycapable}  & 0.246 & 150.3 & 0.098 \\
FLUX~\cite{labs2025flux1kontextflowmatching}  & 0.276 & 169.9 & 0.112 \\
\bottomrule
\end{tabular}
\vspace{-4mm}
\end{table}
\section{Conclusion}
We introduced \ourmethod, a pose-free, feed-forward framework that unifies reconstruction and instruction-driven editing of 3D scenes from unposed, instruction-edited images. By directly predicting edited 3D Gaussian splats in a single pass—without test-time optimization or pose estimation—our approach departs from the conventional reconstruct-edit-refit pipeline and enables fast, photorealistic, and instruction-aligned rendering. Two design choices are key to making this practical at scale: (i) a SAM2-based recoloring strategy that provides reliable, cross-view-consistent supervision despite the lack of true multi-view edited ground truth; and (ii) an asymmetric input scheme that pairs an edited reference view with unedited auxiliary views, encouraging the network to fuse disparate observations while preserving scene structure. Coupled with lightweight 3D regularization on Gaussian centers, these components yield robust multi-view coherence even when the input edits are imperfect or inconsistent.

To enable fair and reproducible comparisons, we proposed {\ourbench}, a scene-level benchmark spanning 20 diverse scenes and four edit types (Add / Remove / Modify / Global). On this benchmark, {\ourmethod} achieves stronger text-image alignment and improved 3D consistency than recent optimization-based and feed-forward baselines, while being orders of magnitude faster at inference. Ablations confirm the necessity of view-consistent recoloring, asymmetric inputs, and 3D geometric losses, and demonstrate that our method is editor-agnostic and generalizing to a range of 2D editing front-ends.

\vspace{1mm}
\noindent\textbf{Limitations and Future Work.}
Although recoloring provides stable supervision, it does not fully capture large geometric changes (\eg, substantial add/remove edits) or extreme material and illumination shifts. Looking ahead, promising directions include expanding supervision beyond recoloring with view-consistent generative augmentation, learning per-object disentangled controls for precise 3D edits, incorporating uncertainty-aware rendering to handle ambiguous or conflicting edits, extending to dynamic scenes and longer view sequences, and further enriching {\ourbench} with more diverse and complex scenes.

{
    \small
    \bibliographystyle{ieeenat_fullname}
    \bibliography{main}
}


\end{document}